\pdfoutput=1

\documentclass[11pt]{article}

\usepackage[final]{acl}

\usepackage{times}
\usepackage{latexsym}
\usepackage{tabularx}
\usepackage{booktabs}

\usepackage[T1]{fontenc}

\usepackage[utf8]{inputenc}

\usepackage{microtype}

\usepackage{inconsolata}

\usepackage{graphicx}

\usepackage{booktabs}
\usepackage{multirow}
\usepackage{multicol}

\usepackage{float}

\usepackage{enumitem}
\usepackage{mdframed}
\usepackage{listings}
\title{Computational Analysis of Conversation Dynamics through Participant Responsivity}

\author{Margaret Hughes\hspace{1.25em} Brandon Roy\hspace{1.25em} Elinor Poole-Dayan\hspace{1.25em} Deb Roy\hspace{1.25em} Jad Kabbara \\
        MIT Center for Constructive Communication \\  Massachusetts Institute of Technology \\ \texttt{\{mhughes4, bcroy, elinorpd, jkabbara, dkroy\}@mit.edu}}

\begin{document}
\maketitle

\vspace{-1em}
\begin{figure*}[t]
    \centering
    \includegraphics[width=1\linewidth]{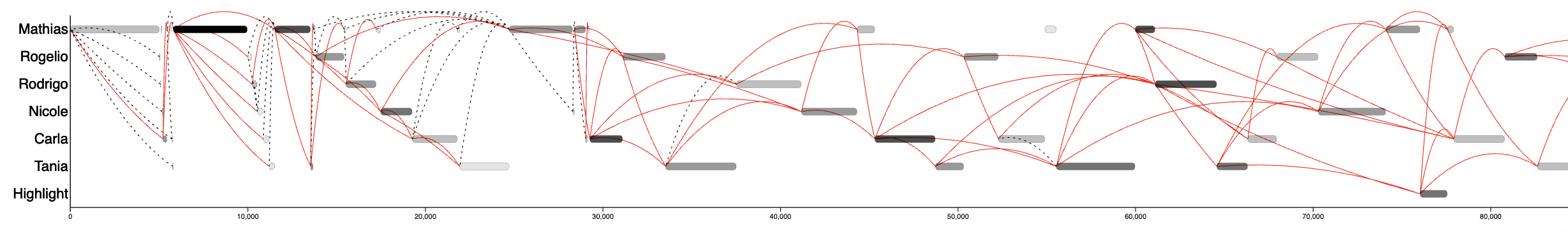}
    \caption{Conversation Map showing the flow of turns, their sequence, and the responsivity links.}
    \label{fig:unfolded}
\end{figure*}

\begin{abstract}
Growing literature explores toxicity and polarization in discourse, with comparatively less work on characterizing what makes dialogue prosocial and constructive. We explore conversational discourse and investigate a method for characterizing its quality built upon the notion of ``responsivity''---whether one person's conversational turn is responding to a preceding turn. We develop and evaluate methods for quantifying responsivity---first through semantic similarity of speaker turns, and second by leveraging state-of-the-art large language models (LLMs) to identify the relation between two speaker turns. We evaluate both methods against a ground truth set of human-annotated conversations. Furthermore, selecting the better performing LLM-based approach, we characterize the nature of the response---whether it responded to that preceding turn in a substantive way or not. 

We view these responsivity links as a fundamental aspect of dialogue but note that conversations can exhibit significantly different responsivity structures. Accordingly, we then develop conversation-level derived metrics to address various aspects of conversational discourse. We use these derived metrics to explore other conversations and show that they support meaningful characterizations and differentiations across a diverse collection of conversations.

\end{abstract}

\section{Introduction}
Trust in government is decreasing rapidly while political polarization increases. The toxicity that is pervasive in social media platforms like Twitter/X has seeped into our engagement offline. Town halls, community forums, and various other means of civic participation have grown hostile and unproductive \cite{Innes_2004, SpeakingPublic}. Democracy scholars call for systems that improve the health of the public sphere and for avenues that enable civic agency and dignity \cite{allen2023justice}. Such systems enable citizens to gather to discuss meaningful ideas, work together to develop a shared understanding, and potentially even reach consensus in decision making processes. One example is citizens assemblies where groups selected through sortition gather together to learn, deliberate, and develop recommendations to their governing body based on the needs and goals of their community through small group, facilitated conversations \cite{chwalisz2019new,chwalisz2020reimagining}. These conversations can surface insights that prove valuable not only as a mirror to one's own community, but also as a portal into the thoughts and needs of a group for leadership or outsiders. 

Growing literature explores toxicity, polarization, and decreased liberties within discourse, and while this understanding is important, that is only half of the challenge. Aspirationally, we strive not just for neutral discourse spaces, but actively constructive, healthy, and rich communication spaces. But how do we evaluate the quality of discourse with respect to these goals? We draw inspiration from collaboration literature and facilitated dialogue practice and argue that within a conversation, one fundamental ingredient for the constructiveness of conversation is {\it responsivity}: the extent to which participants in a dialogue actively listen to, respond to, and build upon one another. To understand this behavior in conversations, we operationalize and evaluate responsivity as a conversation quality metric. In simple terms, responsivity captures whether one person's conversational turn is responding to a preceding turn.

We develop and evaluate methods for quantifying responsivity---first through semantic similarity of speaker turns, and second by leveraging state-of-the-art large language models (LLMs) to compare the relation between two speaker turns. We evaluate both methods against a ground truth set of human-annotated conversations. Selecting the better performing LLM-based approach, we characterize the nature of the response---whether it was responding to that preceding turn in a substantive way or not.

While we view responsivity links as a fundamental aspect of dialogue, we note that conversations can exhibit significantly different responsivity structure. Accordingly, we develop conversation-level derived metrics as a lens through which to examine different aspects of conversational discourse. We use these derived metrics to explore different kinds of conversations and show that they support meaningful characterizations and differentiations (that align with actual differences in purpose, style, etc.) across a diverse collection of conversations. 

\section{Related Work} 

\subsection{Facilitated Dialogue}
Facilitated dialogue is a conversation structure in which a facilitator guides participants in having a conversation according to a pre-designed conversation guide. Facilitators act as neutral stewards of the conversation to ensure conversation norms are respected, intervening in cases of norm violation. In the conversation space we focus on, one additional goal for facilitation is to encourage participants to share personal experiences rather than opinions. The goal of these conversations is thus not to persuade others or win through argumentation, but rather to understand one another more deeply through sharing of personal experiences. This approach is taken within conflict resolution and for community building and can be used to facilitate empathy, understanding, and connection among participants, within and across divides.

Further, facilitated dialogue is a technique used within deliberation and civic discourse spaces such as Citizens' Assemblies \cite{chwalisz2019new,chwalisz2020reimagining} which consist of a random sample of people from a constituency gathered together to deliberate on specific social and political issues. In Citizens' Assemblies and other deliberative spaces, dialogue is the primary means of participation in the public sphere and empowers the various civic actors to practice agency and participate in their governance.

The concept of responsivity builds upon established theoretical frameworks in dialogue studies and communication theory. Drawing from \citet{bakhtin2010dialogic}'s notion of dialogic responsiveness, where meaning emerges through the dynamic interaction between speakers and listeners, our operationalization of responsivity captures the fundamental dialogic principle that authentic communication requires active engagement with others' contributions rather than mere sequential turn-taking.

This theoretical foundation aligns with research in conversational analysis and discourse studies that emphasizes the collaborative nature of meaning-making in dialogue \cite{schegloff2007sequence}. By quantifying the extent to which participants build upon, reflect back, and meaningfully engage with preceding contributions, our metrics operationalize key principles from facilitated dialogue practice where the quality of listening and responding directly impacts the depth and constructiveness of collective understanding.

The distinction between substantive and mechanical responsivity further reflects established practices in dialogue facilitation, where facilitators distinguish between responses that advance collective understanding and those that merely maintain conversational flow without adding substantive content \cite{isaacs1999dialogue}. This theoretical grounding situates our computational approach within broader scholarly understanding of what constitutes effective dialogue and meaningful human interaction.

\subsection{Conversation Metrics}
While tools like 
Jigsaw's Perspective API \cite{lees2022perspective} has been used widely to evaluate dynamics within online conversation spaces \cite{choi2015characterizing,Saveski2021},
rarely are these metrics developed relationally--each comment or Tweet is generally treated independently of those that came before. Interactions and relationships between people are not the unit of analysis, but rather the text in isolation. 

Recent work has started to explore what makes conversations pro-social and constructive. \citet{Bao2021} look at turn-specific measures to identify pro-social conversation dynamics on Reddit. These metrics observe individual conversation contributions rather than the interwoven dynamics that emerge from participants responding to and being in relationship with one another. 

\citet{dowell2019group} develop a system of metrics exploring the relationship between conversation participants.
They present a core metric, namely responsivity, or the tendency of an individual to respond (or not) to the contributions of their collaborative peers. Responsivity is calculated based on the cosine distance between conversation turn embeddings -- the closer the embeddings, the more responsive participants are to one another. We are inspired by this work for our own metric development and evaluation. We believe the choice to center relationship and context when evaluating and describing conversation aligns with the core values and practices in facilitated dialogue. 

Another closely related line of work is conversation disentanglement, which seeks to separate interleaved conversational threads in multi-party dialogue \cite{zhu-etal-2021-findings}. Disentanglement methods aim to recover the underlying “who-responds-to-whom” structure, often in chaotic or asynchronous contexts such as IRC channels \cite{kummerfeld-etal-2019-large}. More recent work has extended this to novel domains such as scripted or dramatic dialogue \cite{chang2023dramatic}. Our approach builds on this tradition of mapping conversational links but diverges in its aims: while disentanglement focuses on thread recovery, we distinguish types of response (mechanical vs substantive) and develop conversation-level metrics to characterize the quality of dialogue. In this sense, our work complements disentanglement by enriching structural mappings with relational measures of responsiveness and constructiveness.

\subsection{LLMs for Social Science}

Recent advances in NLP have enabled more widespread use of LLMs in social science settings. LLMs have been applied to analyze social dynamics in several ways, including understanding emotional undertones \citep{dutt2024leveraging}, social stances in conversations online \citep{chae_davidson_2023}, as well as to extract speaker characteristics \citep{jurafsky-etal-2009-extracting,broniatowski-extracting-2012}.
In particular, leveraging LLMs as zero or few-shot  annotators has been shown to be extremely promising \citep{gilardi-chatgpt-2023,wang-etal-2021-want-reduce, ding-etal-2023-gpt,he-etal-2024-annollm,huang-chatgpt-2023}, potentially even for subjective, nuanced tasks \citep{ziems-etal-2024-large,ruckdeschel-2025-just,xiao-qual-2023}. This may open up NLP research to tackle more complex, interdisciplinary, or niche datasets for which human annotation is very difficult or expensive \citep{ruckdeschel-2025-just}.

However, there is concern that LLMs trained on synthetic data may struggle on highly subjective tasks \citep{li-etal-2023-synthetic}. More work has shown that LLM annotation performance may struggle with conversational data \citep{ziems-etal-2024-large} and that models can be highly variable to prompts \cite{atreja2024promptdesign}. 
Some studies suggest that practitioners should use caution when using LLMs to annotate data \citep{pangakis_automated_2023,huang-chatgpt-2023}.
\citeauthor{pangakis_automated_2023} argue that the use of LLMs to automate annotation for research must always validate performance against human-annotated labels. Motivated by this, our methodology follows the best practices outlined by previous works.

Further, we build upon a growing literature that uses LLMs as a means to annotate and understand discussions at a large scale -- a task previously quite inaccessible. We look to \citet{korre2025evaluation} who do an overarching survey of how LLMs are used to evaluate and facilitate conversation, especially in digital domains like Reddit. We build upon their work by including key features such as turn taking in our analysis, but expand it further to pay particular attention to responsivity, or response relationships between participants, as a critical component of dialogue. Others explore methodologically various means to apply these metrics and evaluate their accuracy through mixed-methods approaches using LLMs to measure constructiveness in conversations, finding LLMs and hybrid-LLM approaches effective for the task \cite{zhou2024llm}. A great deal of work further explores means of evaluation discourse and deliberation through descriptive metrics such as open-mindedness, equality of participation, a general respect for others, or progress towards a common goal \cite{barrett2024beyond, ercan2022research}. Yet, a gap continues to persist around responsiveness and connection to other participants.

\section{Responsivity} \label{sec:resp}

As dialogue is about connection between participants, we explore how participants actively listen to, build upon, and reflect back contributions of those before them through responsivity. While responsivity is not the only metric of importance when understanding conversation dynamics or quality, it is an important and understudied component of conversations, so we start with it in our work. 

In studying responsivity, we define our unit of focus to be a conversation turn. This is the contiguous sequence of utterances a conversation participant speaks until another speaker starts their turn. Previous work  considers responsivity between participants over a whole conversation \cite{dowell2019group}, while we calculate responsivity between participants across windows within a conversation to observe more granular, concurrent interactions. This granularity enables us to not only explore one conversation in summary, but to identify moments of conversations that yielded higher or lower responsivity, or highlighted relationships between participants at key interactions. We further explore the concept of responsivity by distinguishing between two kinds of responsivity:

\begin{itemize}
    \item \textbf{Substantive responsivity:} An interaction where one person meaningfully engages with what another has said. It captures how much a speaker reflects back, builds upon, inquires about, or connects to other ideas, emotions, or experiences shared by the previous speaker, or answers a meaningful question from a previous speaker. 
    \item \textbf{Mechanical responsivity:} An interaction that occurs when a speaker responds in a way that acknowledges or moves the conversation forward but does not add substantial new content. These responses may include polite phrases, conversational hand-offs, or social cues.
\end{itemize}

Responsive structures are integral to many forms of human interaction and conversation. As such, there is a need to better define and identify the boundaries of constructive communication. However, there is no existing method to map a conversation structure to understand how participants respond to and build upon one another. Furthermore, for humans to annotate a conversation for these turns is inefficient, especially if one might want to iterate upon a conversation design based on the responsivity within a previous conversation quickly, or if one would like to understand dynamics in a large corpus. Therefore, we ask how we might automate responsivity annotation. We describe an initial set of automation methods in the following section, along with the methods used to evaluate those approaches. 

\section{Methods}
To automate annotating responsivity, we explore semantic similarity metrics and the use of LLMs via prompting. We develop a crowdsourced human annotation task to evaluate the automated methods and design an interactive data visualization. 

\subsection{Semantic Similarity}\label{sec:semsim}
One approach to operationalizing responsivity is through {\it semantic similarity}, motivated by the idea that the content of the response should have some semantic overlap with the turn to which it is responding. To compute semantic similarity between conversation turns $i$ and $j$, we first obtain the sentence embedding of each conversation turn using MPNet \cite{song2020mpnet}, a deep-learning based embedding model.\footnote{We use all-mpnet-base-v2: \url{https://huggingface.co/sentence-transformers/all-mpnet-base-v2}} We then compute the cosine distance between the embedding vectors for turns $i$ and $j$. For a given turn, we compute its cosine similarity to the preceding 10 turns, and form a responsivity links to those turns with responsivity above a threshold.

\subsection{LLM Approach}
Large language models have shown remarkable performance across many tasks that involve certain kinds of reasoning, content analysis, and the generation and synthesis of text. We hypothesize that an LLM might be able to interpret the meanings in conversation turns in a more nuanced way compared to semantic similarity based on the MPNet embedding model.

We use two state-of-the-art LLMs: GPT-4o \cite{openai_gpt4o_2024} and Claude 3.5 Sonnet \cite{anthropic_introducing_2024}\footnote{We use model versions \texttt{gpt-4o-2024-08-06} and \texttt{claude-3-5-sonnet-20241022} via their respective APIs.} to carry out three increasingly fine-grained tasks, the first of which is equivalent to the semantic similarity task introduced in Section \ref{sec:semsim}.

\textbf{Stage 1 (turn-level linking):} Given a speaker turn and the 10 preceding conversation turns as context, the LLM must identify which (if any) of the preceding turns the current one responds to.

\textbf{Stage 2 (segmentation):}
Given a pair of speaker turns in which one responds to the other, the second stage aims to identify which exact part of the turn responds to which exact part of the given preceding turn. Each (sub)part is called a segment.

\textbf{Stage 3 (classification):} The goal of the third stage is to classify each pair of responsive segments as mechanical or substantive. 

Following best practices for LLM annotation tasks \cite{pangakis_automated_2023}, we perform three runs of the first and third stages for each conversation. To maximize consistency, we drop any labels that appear in less than 2 out of the 3 runs. Full prompts can be found in Appendix~\ref{appendix:prompts}.

\subsection{Human Annotation Task}
In order to better understand how the models structure these conversations, we design and deploy a human annotation task to develop a human set of annotations and links to compare our computationally generated structures. We design the task to be completed on the crowdsourcing research platform Prolific and specify that crowdworkers need to be fluent in English. We pay each Prolific worker \$17 an hour, \$2 above minimum wage in our research institution's state. Each task is to annotate one full conversation, and the worker completion time ranged from 30 minutes to 1.5 hours. When first deployed, we invited workers to give feedback on the complexity of the task, and the feedback we received was that it was a complex task at first, but after a few turns, they were able to complete it with greater ease.  The study was reviewed and approved by the Ethics Review Board.

We had three annotators per conversation, calculated the inter annotator agreement, and treated a majority vote as the human standard to which we compared the LLM and semantic similarity annotations. For the human annotations, we did not ask participants to distinguish between substantive and mechanical responsivity, and hope to evaluate this with human annotation in the future.

\section{Evaluation} 

\subsection{Dataset}
To investigate responsivity as a metric for conversation quality and to evaluate our proposed methods, we use the Fora dataset \cite{schroeder-etal-2024-fora}, a dataset of 262 richly annotated facilitated conversations that were hosted with partner organizations seeking to engage their members and surface insights regarding issues like education, elections, and
public health. Similar to the spirit of our work, the conversations in the Fora dataset center around the sharing of stories and personal experiences, which has been shown \cite{kessler2024hearing} to elicit higher levels of empathy and understanding (compared to the sharing of opinions, especially for polarized sides). The Fora conversations are mostly small-group conversations (median of 6 participants) that were held between 2019-2023, recorded with consent, and automatically transcribed. The organizer was able to redact and edit transcripts for any transcription mistakes following the conversation. 

Alongside this corpus we include a set of conversation collections that are similarly small-group facilitated dialogues that are recorded and transcribed, but that we would expect to be structured differently. Specifically, we included recorded game-play of conversation-based games, a collection of facilitated deliberation sessions from a citizen's assembly developing policy recommendations, and recorded conversations of youth discussing themes in a youth-focused documentary. While these collections contain many similar core attributes, we anticipate them to be recognizably different based on our derived metric set.

\subsection{Responsivity Annotation Evaluation} 
\begin{figure}
    \centering
    \includegraphics[width=\columnwidth]{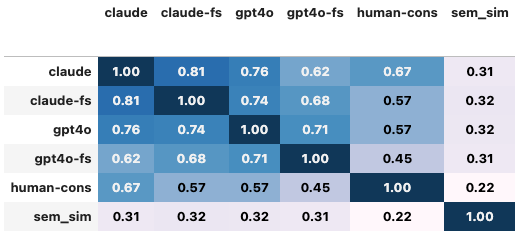}
    \caption{Inter-annotator agreement matrix between annotation methods. Note: fs refers to few shot and cons refers to consolidated.}
    \label{fig:confusion}
\end{figure}

We evaluate differences between annotations using the Jaccard index. The Jaccard index, also known as the Jaccard similarity coefficient, quantifies the overlap of two sets as the size of their intersection divided by the size of their union. For our purposes, when two annotators provide identical annotations on a conversation turn, the Jaccard index will be 1, while a completely non-overlapping set of annotations yields a Jaccard index of 0.

Across the data, the average number of responses per turn annotated by GPT-4 across two conversations is 1.42. For Claude, 1.25. for humans, 1.04. Of those annotations, the average number of substantive responses per speaker turn is 0.99 for GPT-4 and 0.78 for Claude. 
Within our annotated dataset, two conversations were a part of the Fora corpus's story and personal experience annotation scheme. Using those previously annotated conversation turns, we see that $29\%$ of speaker turns are labeled as sharing a personal story/experience. For instances of personal stories and/or experiences, we observe a much higher responsivity rate to those speaker turns, and specifically a higher rate of \textit{substantive} responses. Specifically, non-personal story contributions that were responded to received mechanical responses about 40\% of the time, while story contributions that were responded to received mechanical responses about 30\% of the time.

\begin{figure}
    \centering
    \includegraphics[width=0.7\linewidth]{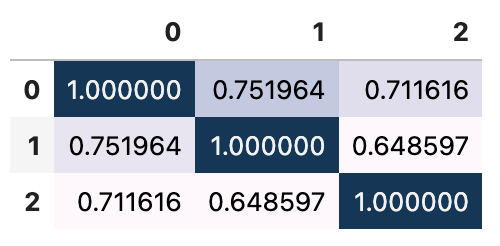}
    \caption{Human inter-annotator agreement matrix for conversation 1113.}
    \label{fig:human-confusion}
\end{figure}

The inter-annotator agreement matrix in Figure~\ref{fig:confusion} shows that Claude and GPT-4 were most aligned out of the annotation methods, followed by Claude and human annotation. The least aligned method of responsivity mapping is the semantic similarity approach, with no Jaccard index greater 0.32. Further, we see that inter-annotator agreement calculated through the Jaccard index for human annotation is low, with an average Jaccard index across all conversations of 0.592. One can see the inter-annotator agreement matrix for a single conversation visualized in Figure~\ref{fig:human-confusion}, showing alignment on par with the LLM to human alignment.  Interestingly, the few shot LLM prompts generally yielded lower inter-annotator agreements than the one-shot methods. In Appendix \ref{appendix:disagreement}, we highlight moments of disagreement between human annotators and LLMs to exemplify how interpretations of responsivity depend on one's position, but the breadth of interpretations is reasonably bounded.

We then compared accuracy of LLM annotations against human annotations for substantive versus mechanical labels with a subset of 100 conversation snippets. Using the preceding context window of “possible responsivity links” for each turn, there was agreement between human and LLM annotators on 91.9\% of the labels (10.6\% links present for both, 81.3\% absent for both). In cases where they disagreed, humans exclusively labeled a link 3.3\% and LLMs 4.8\% of the time. The analysis shows high levels of agreement, and reveals nuances between LLM and human understanding, such as LLMs label responses as substantive slightly more often than humans.

\section{Conversation Analysis}

In the preceding sections, we proposed and evaluated approaches to responsivity annotation. While the semantic similarity based method had low agreement with human annotators, LLM-based methods performed well. A dialogue annotated with such responsivity links supports an examination of conversation structure, both visually (see Figure \ref{fig:unfolded}) and through derived summary metrics. We can see that links between speaker turns show how participants build upon one another, and fragmented sections of conversation show no interconnections. Side conversations are visibly distinct from main conversations in the flow, and highly impactful moments seem visible from their many interconnections.

In this section, we describe some conversation metrics that we show support characterization and differentiation of conversations. The first set of metrics derives directly from turn information, while the second set builds upon the responsivity annotations.

The first set includes simple measures such as the number of speakers, the total number of turns and total duration of the conversation, the number of facilitator turns and speaking time, and the corresponding percentages, and the variance in the number of turns across speakers. We also compute distributional features -- namely, the Gini coefficient \cite{gini-dorfman,gini-farris} of both the speaking time and number of turns to quantify how balanced (Gini coefficient near 0) or unbalanced (Gini coefficient near 1) these quantities are across speakers. Finally, we compute the conditional entropy on the speaker turn {\it sequence} to characterize the variability in speaker turn-taking. A perfectly consistent speaking order would yield a conditional entropy of 0, increasing to a maximum for a random speaker turn ordering.

Since a conversation is not simply a sequence of turns, but rather a sequence {\it connected} by participants responding to one another, we develop additional metrics to characterize this responsivity structure.

The simplest metrics are the rates of substantive and mechanical responsivity. However, since the preceding window used for responsivity annotations may include the speaker as well as the facilitator, we also calculate rates restricting to the subset of non-self and non-facilitator turns. As above, we calculate distributional metrics, but here quantify how actual {\it responses} are distributed across participants using the Gini coefficient. We compute the Gini coefficient on the distribution of substantive responsivity to quantify whether everyone was equally substantively responsive or whether it was concentrated on only a few participants (we also compute variations on the subset of preceding turns considered to exclude facilitator and self). Finally, we compute the conditional entropy of the distribution of who substantively responds to whom, helping quantify whether everyone generally responded to everyone else or whether participants were more selective in who they responded to. See Appendix \ref{apdx:features}, Table \ref{tab:feature-definitions} for a summary description of all features.

The conversation-level metrics support comparing between conversations and analysis of large conversation collections. For this evaluation, we analyzed conversations in the Fora Corpus as well as the youth documentary discussions (n=11), the citizen assembly (n=13), and game-play data (n=12). As described above, some of these collections have different purposes and formats.

To begin, we computed all features (23 in total) for each of the 101 conversations described in Appendix \ref{apdx:features}. The features themselves are motivated by observations and experience with small-group dialogue. Since we know that some of these features are correlated (see Appendix \ref{appendix:earlier-ver}, Figure \ref{fig:heatmap-all-feat-corr}), so to support interpretability we identify groups of highly correlated features and take only a subset. We did this manually given the small number of features and our original goal of capturing certain aspects of conversation structure. For example, Figure \ref{fig:heatmap-all-feat-corr} shows that \texttt{avg\_subst\_responded\_rate} and a block of functionally related (but more specific) features are highly correlated. In this case, we decided to keep the base feature and the related ``non-self'' feature, since we felt it reflected an important distinction (preferring responses to others rather than oneself). We note that we did experiment with PCA on the original feature set, finding that only 9-10 features are needed to preserve 95\% of the variance. However, the resultant components are much harder to interpret as they are linear combinations of the original 23 features. 

This process yielded 12 features from the original 23, listed in Figure \ref{fig:feature-by-cluster-heatmap}. 
These include both direct features (e.g. speaking time, percentages, etc.) as well as features derived from responsivity annotations (e.g. substantive and mechanical responsivity rates, entropy, etc.). 
We clustered the conversations using these features, first by applying UMAP \cite{McInnes2018} to reduce the features to 3 dimensions, and then clustering with HDBscan \cite{McInnes2017}. This yielded 5 clusters, which we describe below. Further, we visualize the conversations in a 2-dimensional UMAP-reduced cluster-colored plot, in Figure \ref{fig:umap_five}.

\section{Conversation Clusters Analysis Results}
\begin{figure}
    \centering
    \includegraphics[width=1\linewidth]{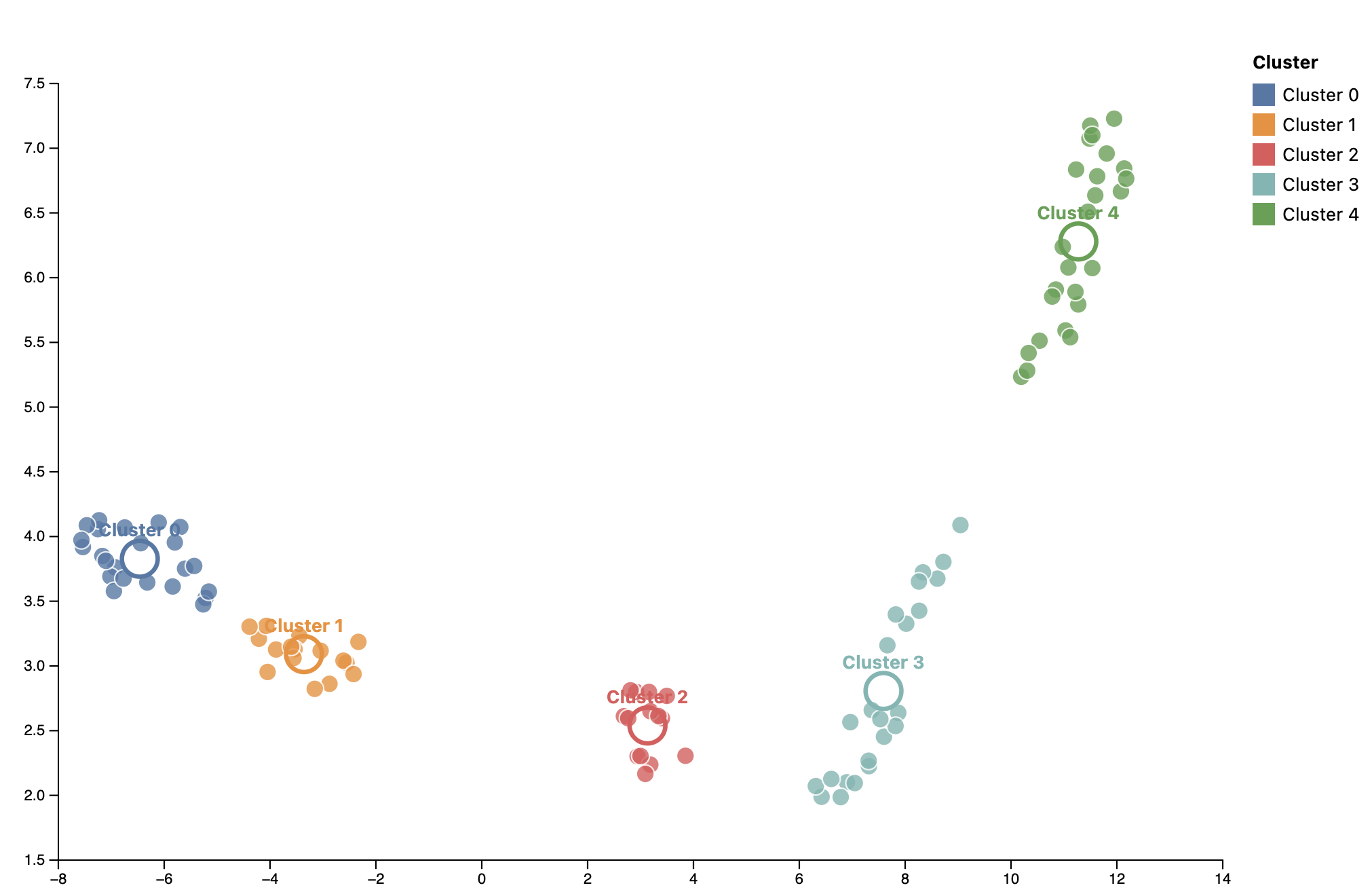}
    \caption{Cluster Map showing the 5 clusters.}
    \label{fig:umap_five}
\end{figure}
\begin{figure}
    \centering
    \includegraphics[width=1\linewidth]{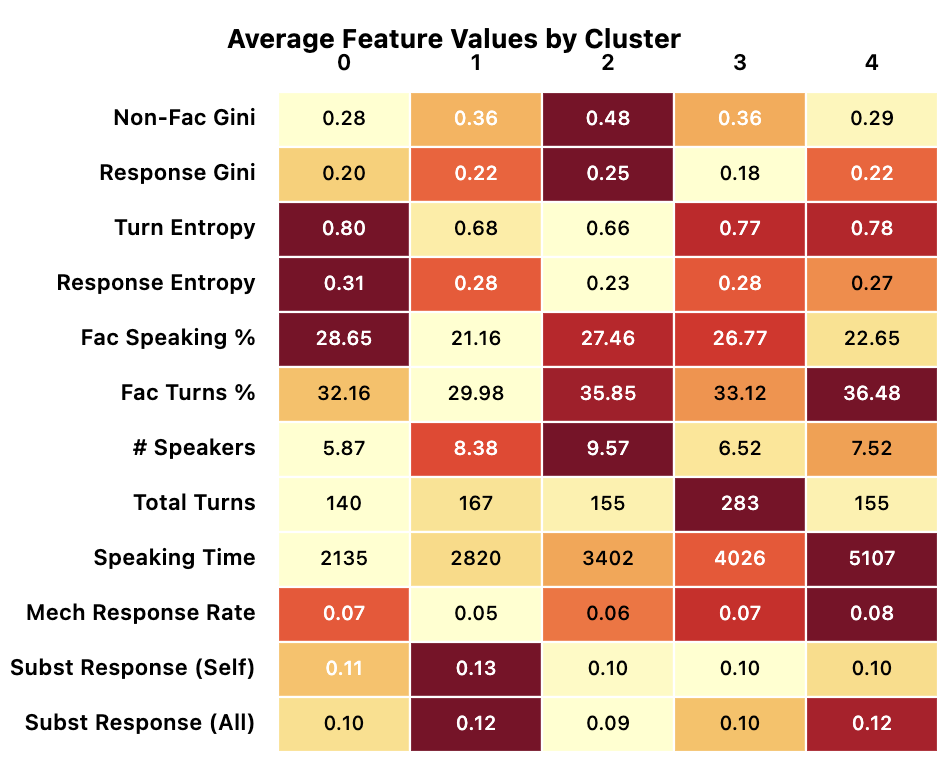}
    \caption{Heatmap showing the average feature values by cluster.}
    \label{fig:feature-by-cluster-heatmap}
\end{figure}

In the following section, we describe the clusters identified through our cluster analysis to accompany the centroids outlined in Figures \ref{fig:umap_five} and \ref{fig:feature-by-cluster-heatmap}.

\textbf{0: Facilitated, Dynamic Small Groups.}
These conversations are shorter and involve smaller groups. They are marked by high equality in speaking time and high entropy in both turn-taking and responsivity, suggesting a free-flowing, dynamic exchange. Facilitators play a relatively prominent role in speaking, while substantive responsivity is at a moderate level.

\textbf{1: Participant-Driven, Substantively Engaged Dialogues.}
This cluster has the highest rates of substantive responsivity, and the lowest rates of mechanical responsivity. With the lowest percentage of facilitator speaking time, these show participant driven, responsive conversations. This cluster holds mid-levels of entropy and gini coefficient, showing a balanced but structured flow. 

\textbf{2: Structured, Unequal, Large-Group.}
Conversations in this cluster are the largest and most unequal in terms of participation. They show the highest Gini coefficients and the lowest entropy, pointing to a highly structured and facilitator-dominated format. Substantive responsivity is the lowest, suggesting a less engaged or more top-down dynamic.

\textbf{3: High Turns, Disordered, Low Response.}
These are long conversations with a high number of turns and slightly elevated entropy, implying less orderly interaction. Substantive responsivity is low, suggesting limited depth or follow-through in exchanges. Average speaking time time is modestly elevated, mechanical responsivity is higher and substantive is lower. That, tied with a slightly elevated facilitator speaking time suggests longer, disordered, less substantive, more facilitator driven conversation. This pattern often correlated with face-paced conversation games.

\textbf{4: Inclusive, High-Engagement, Long.}
This cluster includes the longest conversations by speaking time and the highest percentage of facilitator turns though facilitator speaking time is quite low, suggesting they take many brief turns. Conversely, participant take few turns, but have the longest speaking time, suggesting long, extensive turns. It features the lowest speaking inequality, indicating a highly inclusive dynamic, and notably high substantive responsivity when self and facilitator contributions are excluded. The slightly elevated responsivity Gini suggests a less predictable, possibly more exploratory style of engagement.

\section{Discussion}

In this work, we argue that understanding constructive conversation requires first examining the underlying structure of contributions and responses. How participants listen, respond, and build upon one another is foundational to conversation, shaping its flow, emergent relationships, and overall quality. We build on prior work introducing responsivity as a key metric, applying it across conversation windows rather than as a summary statistic.

Expanding on the approach of \citet{dowell2019group}, we prompt state-of-the-art LLMs to annotate conversations for responsivity and, for the first time, evaluate this method against human annotators. Our findings indicate that LLMs align more closely with human judgments than semantic similarity-based approaches. However, disagreement among human annotators highlights the inherent difficulty and subjectivity of the task. Notably, while LLMs do not perfectly match human annotations, inter-LLM-alignment is comparable to variations observed among humans

Using our approach, we further observe responsivity dynamics validating our expectations from the dialogue literature.

For example, in facilitated discussions, participants are expected to be more responsive to one another than to facilitators—a pattern reflected in our responsivity data \cite{wilson2002circles, schroeder-etal-2024-fora}. Likewise, sharing stories tends to foster stronger connections between participants than opinions, as signaled by higher responsivity, particularly substantive responsivity, to story-based contributions.

Building on these observations, we introduce a set of derived metrics designed to capture meaningful distinctions in conversation styles and structures. Applying these metrics to a diverse set of conversations, we demonstrate that clustering produces interpretable groupings that align with actual differences in conversational purpose, style, and structure. These findings suggest that our derived metrics effectively distinguish between conversation dynamics, further validating their utility in dialogue analysis.

The distinction between substantive and mechanical responsivity proves particularly valuable in characterizing conversation quality. Our analysis reveals that conversations with higher rates of substantive responsivity tend to exhibit different structural patterns than those dominated by mechanical responses. This finding supports theoretical frameworks from dialogue studies that emphasize meaningful engagement over mere acknowledgment as a marker of constructive interaction.

Furthermore, our conversation-level clustering analysis demonstrates that responsivity patterns can meaningfully differentiate between conversation types. The five clusters we identified—ranging from “Facilitated, Dynamic Small Groups” to “Inclusive, High-Engagement, Long” conversations—each exhibit distinct responsivity signatures that align with their intended purposes and facilitation styles. This suggests that responsivity metrics capture not just individual interaction quality but also systemic properties of different conversational contexts.

\subsection{Applications and Integration Opportunities}
We see three primary implications of this work. First, conversation has long been and will continue to be a key medium for democratic civic participation. While extensive research examines non-constructive, toxic, and polarizing discourse, existing work on constructive communication has primarily focused on dyadic relationships \cite{gable2018you, rusbult1991accommodation}, educational settings through collaborative learning research \cite{johnson1999making}, and social-emotional learning frameworks \cite{weissberg2015social}. However, less attention has been given to systematically measuring and characterizing constructive communication patterns in multi-party civic dialogue contexts. Our work provides a method to support those designing constructive communication spaces by helping them assess and refine their interventions. This kind of reflection, aimed at improving conversational dynamics, builds on existing work such as Meeting Mediator and Keeper \cite{kim2008meeting, hughes2021keeper, Adachi}.

Second, while we do not apply our approach to online conversations in this work, we believe it holds significant potential for digital spaces. Comment sections of videos and news articles, subreddits, or threads on microblogging platforms could benefit from analyzing not only explicit replies but also responses that build upon previous contributions. This shift would allow us to examine both the structured conversational elements embedded in platform design and the more nuanced interactions experienced by participants—structures often invisible in traditional data analysis.

Finally, as more governance processes and public discourse move online through tools like Pol.is \citep{small_polis_2021} and Remesh \citep{konya_democratic_2023}, evaluating online conversations will become increasingly important. Digital communities require governance and moderation, and we believe this approach could support prosocial moderation by providing insights into the underlying conversational dynamics of online communities.

\subsection{Future Work}
While responsivity is a foundational metric, we do not see it as the sole indicator of constructive conversation. Drawing from both theory and practice, we recognize that elements such as personal story sharing, introducing new ideas, and other core aspects of dialogue also signal value and connection. We aim to expand our framework to incorporate a more comprehensive set of metrics in future work. With this expanded set of metrics, we also hope to build on Dowell’s work by developing a taxonomy of participation types. If we cluster participants based on these characteristics, do clear patterns emerge? Finally, we seek to apply these metrics across “high” and “low quality” conversational types. Can this approach effectively analyze different conversation quality? Can it help identify constructive communication across varied contexts? And does it enhance our understanding of prosocial and constructive discourse in both digital and in-person spaces?

\section{Conclusion}
In this work, we map responsivity between conversation participants in multiple ways to reveal conversation structures. This work, we believe, can help lay the foundation for developing a more comprehensive set of conversation metrics so we may understand what makes conversations constructive and healthy. We compare human annotation, a semantic similarity approach, and a large language model approach to describe the alignment between these methods and various conversation structures, such as facilitator acts and storytelling. We then develop a set of metrics derived from the responsivity structure. Clustering and analysis on these derived metrics reveal qualitatively different kinds of conversations, which we find generally align with real differences in purpose and style of facilitated conversations. Looking forward, we believe this work will contribute meaningfully to conversation analysis in various domains within and beyond computational linguistics, ranging from the civic world, to computer-supported cooperative work, to online conversation spaces.

\section{Limitations}
There are several key limitations to this study. First, we have not systematically compared or evaluated the differences between the LLM, semantic similarity, and human annotations. While we understand to what degree they are aligned or not, we have not fully examined the points where they disagree. For example, are there certain kinds of turns that Claude appropriately annotates, that GPT and semantic similarity miss? We could discern this through further qualitative investigation into the disagreement points. Further, while we did collect human annotations, there was meaningful disagreement between human annotators. We are not certain why there was disagreement, on what kinds of speaker turns they disagreed, and if there were meaningful patterns across participants in their annotation styles. Finally, while LLMs show promising alignment with human annotations, they may still encode biases inherent in their training data. Future work could explore these potential biases within the conversation quality and dynamics context. We again hope to improve this study for future work by understanding more deeply these patterns through systematic, qualitative review. With respect to derived metrics, there are surely other metrics that can meaningfully characterize different aspects of conversation structure and dynamics. We have pursued this primarily in an unsupervised setting, but appropriate conversation labels could support determining which features are most salient or informative for the task.

\bibliography{main}

\newpage
\appendix
\onecolumn

\section{Full List of Features with Definitions}\label{apdx:features}
\begin{table}[ht]
\scriptsize
\centering
\begin{tabularx}{\textwidth}{p{0.4\textwidth} X}
\toprule
\textbf{Feature} & \textbf{Definition} \\
\midrule
\texttt{speaking\_time\_gini\_coefficient} & Measures inequality in total speaking time across all participants. Higher values indicate more unequal participation. \\
\texttt{turn\_distribution\_gini\_coefficient} & Measures inequality in the number of turns taken by each participant. \\
\texttt{non\_facilitator\_speaking\_gini\_coefficient}* & Speaking time inequality among non-facilitator participants only. \\
\texttt{non\_facilitator\_turn\_gini\_coefficient} & Turn-taking inequality among non-facilitator participants. \\
\texttt{gini\_subst\_responded\_rate\_nonself}* & Inequality in substantive response rates, excluding self-responses. \\
\texttt{gini\_subst\_responded\_rate\_nonself\_nonfac} & Substantive response rate inequality, excluding self- and facilitator-directed responses. \\
\texttt{gini\_subst\_responded\_rate\_nonself\_exclfac} & Substantive response rate inequality, excluding responses from the facilitator. \\
\texttt{gini\_subst\_responded\_rate\_nonself\_nonfac\_exclfac} & Most restrictive: excludes responses to self and facilitator, and includes only non-facilitator targets. \\
\texttt{turn\_sequence\_entropy}* & Entropy of the speaker turn sequence. Higher values suggest less predictable (more disordered) turn-taking. \\
\texttt{substantive\_responsivity\_entropy}* & Entropy of how substantive responses are distributed across speakers. \\
\texttt{facilitator\_speaking\_percentage}* & Percentage of total speaking time contributed by the facilitator. \\
\texttt{facilitator\_turns\_percentage}* & Percentage of turns taken by the facilitator. \\
\texttt{num\_turns\_facilitator} & Raw count of turns taken by the facilitator. \\
\texttt{num\_observed\_speakers}* & Number of unique speakers in the conversation. \\
\texttt{total\_turns\_in\_conversation} & Total number of turns across all speakers. \\
\texttt{total\_speaking\_time\_seconds}* & Total amount of speaking time in seconds. \\
\texttt{turn\_count\_variance} & Variance of distribution of number of turns across participants. \\
\texttt{avg\_subst\_responded\_rate} & Average rate at which participants give substantive responses. \\
\texttt{avg\_mech\_responded\_rate}* & Average rate at which participants give mechanical responses. \\
\texttt{avg\_subst\_responded\_rate\_nonself}* & Substantive response rate directed at others (excluding self-responses). \\
\texttt{avg\_subst\_responded\_rate\_nonfac} & Substantive responses excluding facilitator as response reciever. \\
\texttt{avg\_subst\_responded\_rate\_nonself\_exclfac} & Substantive responses excluding self and directed at others, excluding the facilitator. \\
\texttt{avg\_subst\_responded\_rate\_nonself\_nonfac\_exclfac}* & Most restrictive substantive response average — excludes both self and facilitator, as responder or target. \\
\bottomrule
\end{tabularx}
\caption{Definitions of conversation analysis features. Features used in the final, reduced set have an astrisk.}
\label{tab:feature-definitions}
\end{table}

\section{Initial Clusters}
\label{appendix:earlier-ver}

In an earlier version of conversation clustering, we used the full set of 23 features.

\begin{figure}
    \centering
    \includegraphics[width=1\linewidth]{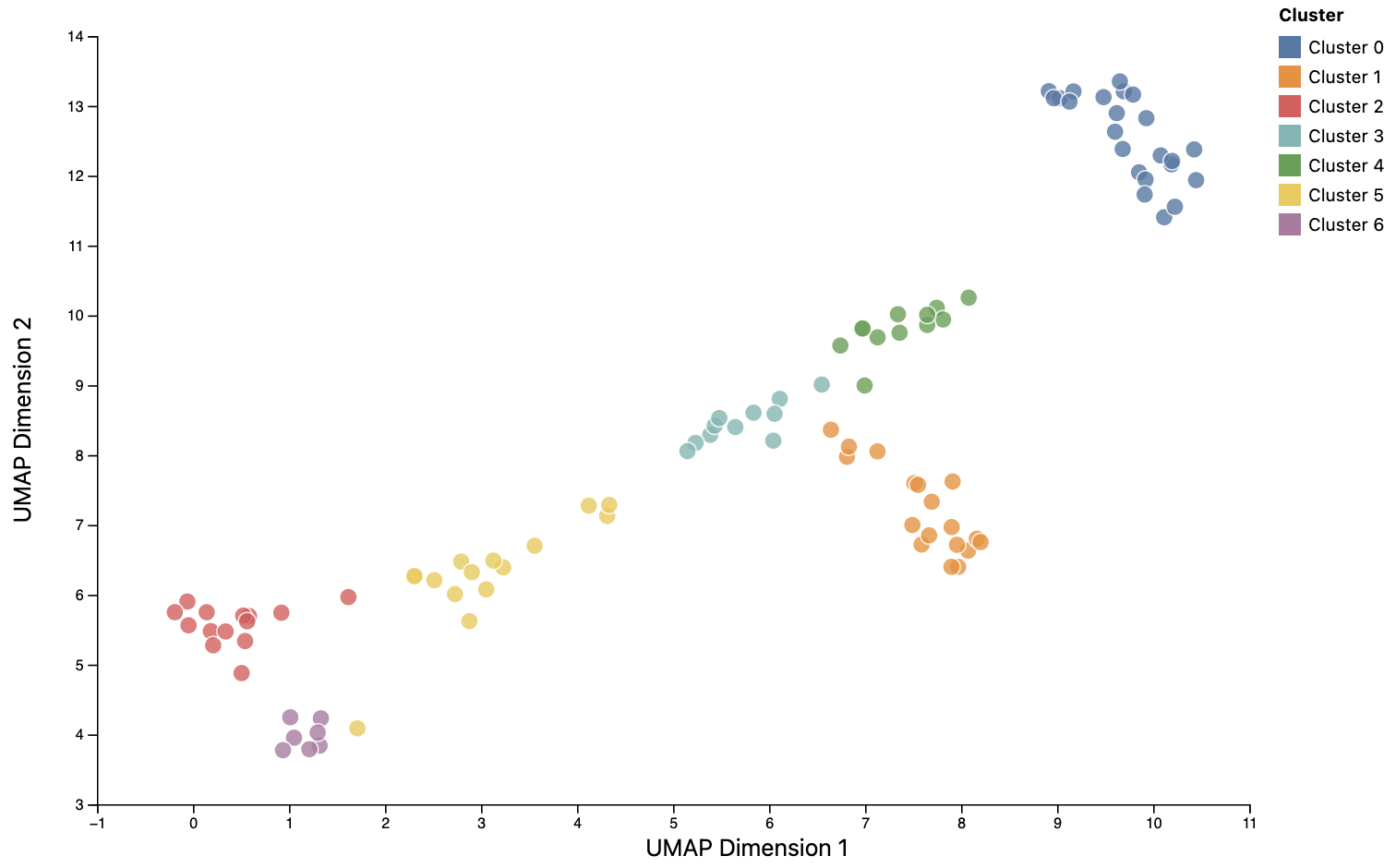}
    \caption{Cluster Map showing the 7 clusters.}
    \label{fig:umap-clustered-orig}
\end{figure}

We applied UMAP \cite{McInnes2018} to these features, reducing to 5 dimensions, followed by clustering with HDBscan \cite{McInnes2017}. For visual inspection, we also applied UMAP to the full feature set to obtain a 2-dimensional visualization, shown in Figure \ref{fig:umap-clustered-orig}. The characteristics (i.e. average feature values) are provided in Table \ref{tab:clusters}. We named the clusters based on their feature characteristics, and describe the clusters below.

\begin{table*}[ht]
\centering
\resizebox{\textwidth}{!}{%
\begin{tabular}{lccccccc}
\toprule
\textbf{Feature} & \textbf{Cluster 0} & \textbf{Cluster 1} & \textbf{Cluster 2} & \textbf{Cluster 3} & \textbf{Cluster 4} & \textbf{Cluster 5} & \textbf{Cluster 6} \\
\midrule
Group Size & 23 & 18 & 14 & 11 & 12 & 15 & 7 \\
Speaking Time Gini & 0.303 & 0.412 & 0.297 & 0.452 & 0.388 & 0.370 & 0.329 \\
Turn Sequence Entropy & 0.775 & 0.780 & 0.803 & 0.653 & 0.743 & 0.697 & 0.781 \\
Substantive Responsivity Entropy & 0.277 & 0.257 & 0.310 & 0.254 & 0.278 & 0.304 & 0.321 \\
Facilitator Speaking \% & 22.293 & 31.070 & 28.356 & 24.806 & 22.963 & 19.467 & 33.684 \\
Avg Subst Responded Rate & 0.122* & 0.075 & 0.143* & 0.101 & 0.106 & 0.116 & 0.071 \\
Total Turns & 137.696 & 426.444* & 103.500 & 113.818 & 107.417 & 127.267 & 227.286 \\
\bottomrule
\end{tabular}%
}
\vspace{1em}
\caption{Clusters from initial clustering on all features, with average feature values for selecte features. Asterisks (*) indicate values deviating more than 1 standard deviation from the global mean.}
\label{tab:clusters}
\end{table*}

\textbf{Dialogue Cluster: }Cluster 0 has medium to low Gini levels for speaking time, slightly elevated turn sequence entropy and facilitator speaking turns percentage, and about average Gini coefficients for responsivity. This cluster contains the majority of the Fora corpus, facilitated dialogues where the facilitator holds the space, takes many turns, and ensures a balance of speaking opportunities.
Further, most of these conversations are on Zoom, consistent with increased facilitator speaking turns percentage.

Cluster 4 exhibits some similar characteristics, though with higher speaking time and turn distribution Gini coefficients. This holds the second greatest number of Fora conversations and repeated facilitators distributed across clusters 0 and 4. For these reasons, we define this cluster set as being the most dialogue-oriented clusters.

\textbf{Dynamic Games: }Cluster 1 has very low substantive responsivity rates, medium levels of responsivity Gini coefficients, and medium to high speaking time Gini coefficients (both with and excluding facilitator turns), with by far the highest total turn counts. We see this cluster contains all game-based, in-person conversations with rapid turn-taking, unequal speaking time, and low responsivity. The games are split perfectly across this cluster and Cluster 6. Cluster 6 has the lowest responsivity Gini rates, the lowest average substantive or mechanical responsivity, and the highest turn-sequence and responsivity entropy, as well as the highest percentage of facilitator speaking time. The game played in this cluster is meant to mimic dialogue and has more structured turn-taking and higher responsivity entropy, showing the few but significant distinctions between the clusters.
\begin{figure}
    \centering
    \includegraphics[width=0.7\linewidth]{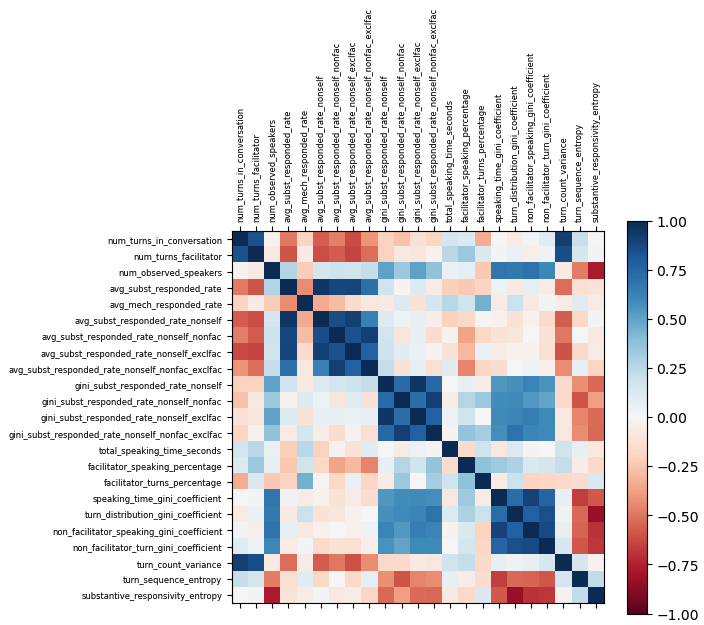}
    \caption{Heatmap showing the correlation between all 23 features. Note the block structure of both positively and negatively correlated features, primarily corresponding to variations on a fundamental measurement (e.g. variations on the speaking time and turn distribution Gini coefficient.)}
    \label{fig:heatmap-all-feat-corr}
\end{figure}

\textbf{Responsive and Balanced: }Cluster 2 has the lowest turn and speaking time Gini coefficients, as well as low average Gini responsivity coefficients. It does exhibit high overall responsivity and the highest turn-sequence entropy. This suggests a highly balanced and responsive conversation, with a well mixed turn ordering between participants. This cluster contains second most conversations from the student assembly and conversations from the youth documentary and could be described as natural feeling and rich.

\textbf{Unequal and Predictable: }Cluster 3 has the highest Gini coefficients with the most unequal speaking time and rates of responsivity, along with the lowest turn sequence and responsivity entropy, suggesting they are highly unequal and very regimented speaking order. On average, they have middle rates of responsivity overall. This cluster contains one facilitator from the youth project and two groups of conversations with one or two individuals who often spoke extensively without the others contributing to the same degree. Based on these characteristics, we describe this cluster as one with unequal contributions, but a highly predictable structure. 

\textbf{Deliberation and Discussion: }Cluster 5 has the highest responsivity rates, the lowest rates of facilitator speaking and turn-taking percentages, and average speaking time and responsivity Gini rates. This cluster contained almost all the conversations with youth discussing the documentary and the student assembly conversations. Their characteristics suggest debate, deliberation, and low facilitator intervention.

\section{Inter-Annotator Disagreement Examples}\label{appendix:disagreement}
To unpack inter-annotator agreement scores, consider this excerpt from conversation 1093 (Table \ref{tab:transcript_excerpt}):

\begin{table}[ht]
\centering
\small
\begin{tabularx}{\textwidth}{clX}
\toprule
\textbf{Turn} & \textbf{Speaker} & \textbf{Utterance} \\
\midrule
20 & Justyce & I will also say mine just so you can get a chance to know who I am. My name is Justyce, I am a student at UW Madison, and I'm a senior and I'm a local voices network intern. My pronouns are she, her, hers. And a value that is important to me is respect. I feel like we learned from a very young age that you should treat others the way you'd like to be treated. So that is a big aspect. Now, I'd like to invite you to share a little about your background. Take a minute and think about a personal story from your life that has shaped who you are. This could be from your childhood, adulthood, or work life, or something you saw happen to someone else. And we can go backwards this time. So Elizabeth, would you like to go first? \\
21 & Elizabeth & Can I have another minute to think? \\
22 & Justyce & Yeah. Meghan, are you ready at all? \\
23 & Meghan & By life event, do you mean a memory that we hold closely to ourselves or...? \\
24 & Justyce & Yeah, that works too. Just something about your background. It doesn't have to be anything specific. And if it's a memory, that works too. \\
25 & Meghan & Okay. I guess I'm interested in diversity, equity, and inclusion because as a woman in STEM, it's hard to be heard compared to other people. So I value intersectional environmentalism, and also recognize that microaggressions are still prevalent in today's society. \\
26 & Justyce & All right, thank you. And Amelia, are you able to go? \\
27 & Amelia & Yeah. So I guess what I wanted to share was that I'm adopted from China. My parents are Caucasian, but I'm Asian. That played a big role in my upbringing—trying to be comfortable with that. It’s kind of weird seeing my family in public, since we don’t all look the same. That taught me to be more accepting of other people. \\
28 & Justyce & Yeah. I definitely know how you feel there. Well, not the adoption part, but being mixed race and along those lines. And Maggie, would you like to go? \\
29 & Maggie & Yeah. I grew up Catholic, which I know is one of the major religions in our society, but the town I grew up in was pretty Jewish. I don’t know the exact proportion, but I... \\
\bottomrule
\end{tabularx}
\caption{Transcript excerpt of participants sharing background stories and values.}
\label{tab:transcript_excerpt}
\end{table}

Consider turns 20-29 – GPT4o and Claude labels turn 29 as a response to turn 20, while semantic similarity labels it as responsive to turn 27, and the human annotator “majority vote” labels it a response to turn 28, with an overall net effect of reducing the annotation agreement (Jaccard) scores between human and machine. In fact, the 6 individual human annotators (note: most conversations annotated by 3 annotators) labeled this turn as follows: a0: [28], a1: [20, 28], a2: [28], a3: [28], a4: [20, 24, 28], a5: [27, 28], netting as 6 votes for turn 28, 2 votes for turn 20, and 1 vote for turns 24 and 27. The somewhat conservative “majority vote” strategy that we used takes only those labels submitted by at least half of the annotators, yielding only label 28 for this turn.

In reading the transcript, it is clear that turn 29 is a response to both the preceding turn (turn 28) and the question by the facilitator (turn 20), as provided by the LLMs. First, in turn 28 Justyce explicitly asks Maggie to share, which she does in turn 29. But in fact, the content of Maggie’s turn (turn 29) is addressing the prompt by Justyce in turn 20. We would argue that both of these are valid, and perhaps best would be to label it as both.

An important point to make regarding the individual human annotations for turn 29 above is that they are actually quite consistent and specific – they do not span all possible preceding turns, but are concentrated on only a few. Having multiple humans randomly labeling would have produced lower inter-annotator agreement than what is shown in Figure 3. However, in future work we may wish to introduce additional mechanisms to ensure annotators spend sufficient time reviewing their selections.

We highlight another example in Table \ref{tab:community_transcript}. In this instance, 3 humans annotators annotated speaker turn 22 to be responsive to the following turns: 16, [16, 19], [19, 20]. GPT annotated it as responsive to the following turns [19, 20]. This case is slightly more blurry. For example, turn 16 is where the facilitator, Fiona, prompts the group with a question. While the first two annotators perceive Kristel's turn as responding directly to that initial prompt, two annotators perceive her as building upon and responding to herself and Libby who spoke right after. Again, no answer is objectively wrong. Turn 22 does seem to build upon the initial contribution from Kristel, and then further on Libby's comments, and may not obviously respond directly to the initial prompt itself, but it does add greater detail to the answer to the prompt and technically responds to it. We hope this example further highlights both the ambiguity of the task, but also the breadth of interpretations. While there are multiple interpretations, none are very far from one another, and out of 10 possible links, simple disagreements emerge around one or two of those links. Again, we show that annotations are quite consistent, but difference can emerge around the nuances of responsivity. 

\begin{table}[H]
\centering
\small
\begin{tabularx}{\textwidth}{clX}
\toprule
\textbf{Turn} & \textbf{Speaker} & \textbf{Utterance} \\
\midrule
16 & Fiona & Perfect, thank you. Our second question is if you could share in a few words what you feel makes a thriving and resilient community? \\
17 & Kristel & That's hard to put in a few words. \\
18 & Fiona & You can use more than a few words. We have a couple people here. So you have, there's plenty of time. \\
19 & Kristel & There's so many angles to approach that from. I could argue that communication is a huge part of that. I think access to resources is our absolute bottom line for that. Access to resources so that we're not living in food insecurity—especially with Maine and Kennebec County specifically having such high levels of food insecurity—that impacts every other area of someone's life. It's not just about feeling hungry. That’s going to impact our young folks' ability to engage in education, which then impacts their future, physical development, and brain development. So something as seemingly simple as food is a building block to getting that thriving, resilient community. \\
20 & Libby & Yeah. And I would just add... Excuse me, I apologize. Something I actually learned from Kristel, who is a great mentor for me. We all play a piece in this puzzle of creating a thriving, resilient community. I think a big piece of that is support, empathy, and understanding that we are all human and doing our best to better ourselves and the environment for others. A big part of our work lately has focused on recognizing strengths and the resources around us, but also understanding that we’re all trying to make it through. Empathy and support are big pieces, along with addressing food insecurity, and making sure people are connected to shelter and food. It's hard to maintain mental health without those needs met. \\
21 & Fiona & Great. \\
22 & Kristel & I tend to go straight into something more tangible, given my background as a social worker. I want to see those pieces meeting a baseline for our folks and then obviously move into higher needs. Just wanted to give context as to why the mental health person went with food. \\
\bottomrule
\end{tabularx}
\caption{Transcript excerpt on defining a thriving and resilient community.}
\label{tab:community_transcript}
\end{table}

\section{LLM Prompts}\label{appendix:prompts}
\subsection{Stage 1 (turn-level linking)}
\begin{mdframed}[frametitle={System Instructions:},
    frametitlefont=\bfseries,
    innertopmargin=0.5em
]
\begin{lstlisting}
Your task is to draw connections between the current, most recent conversation turn and the 
preceding speaker turns in terms of **Responsivity**: the tendency of an individual to respond 
(or not) to the contributions of their collaborative peers.

Now, you will be provided with an excerpt of a conversation, indexed by speaker turn id. Your 
response should be in JSON according to the format specified below.
\end{lstlisting}
\end{mdframed}

\begin{mdframed}[frametitle={Prompt Instructions:},
    frametitlefont=\bfseries,
    innertopmargin=0.5em
]
\begin{lstlisting}
**Conversation excerpt:**
{excerpt}

**Current turn**
{current}

**Output instructions:**
Step 1: Consider the above conversation excerpt.
Step 2: Consider the current turn and whether it responds to any preceding turn.
Step 3: If it does, identify the preceding turn id(s) it specifically responds to in the 
"link_turn_id" field. For not responsive segments, mark ["NA"] in the "link_turn_id" field.

Respond in JSON as follows: 

{{ 
"link_turn_id": List<id of turn(s) responding to if applicable, otherwise ["NA"]>
}}
\end{lstlisting}
\end{mdframed}

\subsection{Stage 2 (segmentation)}
\begin{mdframed}[frametitle={System Instructions:},
    frametitlefont=\bfseries,
    innertopmargin=0.5em
]
\begin{lstlisting}
Your task is to draw connections between two speaker turns in a conversation. Given two speaker turns in which one directly responds to the other, your task is to identify what specific part(s) of the second turn responds to what specific part(s) of the first. 

Your response should be in JSON according to the format specified below.
\end{lstlisting}
\end{mdframed}

\begin{mdframed}[frametitle={Prompt Instructions:},
    frametitlefont=\bfseries,
    innertopmargin=0.5em
]
\begin{lstlisting}
**Speaker Turn 1:**
{speaker_turn_1}

**Speaker Turn 2:**
{speaker_turn_2}

**Output instructions:**
Step 1: Consider the above, in which {speaker_2} responds to {speaker_1}.
Step 2: Identify the part of Speaker Turn 2 that specifically responds to something in the previous turn. This should be an exact quote from Speaker Turn 2.
Step 3:  Identify the part of Speaker Turn 1 that the above is directly responding to. This should be 
an exact quote from Speaker Turn 1.

Respond in JSON as follows: 

{{ 
"step_2": Str<your response to step 2>,
"step_3": Str<your response to step 3>
}}\end{lstlisting}
\end{mdframed}

\subsection{Stage 3 (classification)}
\begin{mdframed}[frametitle={System Instructions:},
    frametitlefont=\bfseries,
    innertopmargin=0.5em
]
\begin{lstlisting}
Your task is to draw connections between two speaker turns in a conversation that respond to each other. Each responsive speaker turn can be either **substantive** or **mechanical**:

- Substantive Responsivity refers to an interaction where one person meaningfully engages with what another has said. It captures how much a speaker reflects back, builds upon, inquires about, or connects to other ideas, emotions, or experiences shared by the previous speaker, or answers a meaningful question from a previous speaker. 

- Mechanical Responsivity, on the other hand, occurs when a speaker responds in a way that acknowledges or moves the conversation forward but does not add substantial new content. These responses may include polite phrases, conversational hand-offs, or social cues.

Your response should be in JSON according to the format specified below.
\end{lstlisting}
\end{mdframed}

\begin{mdframed}[frametitle={Prompt Instructions:},
    frametitlefont=\bfseries,
    innertopmargin=0.5em
]
\begin{lstlisting}
**Speaker Turn 1:**
{speaker_turn_1}

**Speaker Turn 2:**
{speaker_turn_2}

**Output instructions:**
Step 1: Consider the above, in which {speaker_2} responds to {speaker_1}.
Step 2: Determine whether Speaker Turn 2 responds mechanically OR substantively to Speaker Turn 1.
Step 3: If it truly has elements of both mechanical and substantive responsivity, then it should be considered substantive.

Respond in JSON as follows: 

{{ 
"label": Str<"responsive_mechanical", or "responsive_substantive">,
}}
\end{lstlisting}
\end{mdframed}

\section{Annotation Task}\label{appendix:Annotation}
\begin{figure*}
    \centering
    \includegraphics[width=1\linewidth]{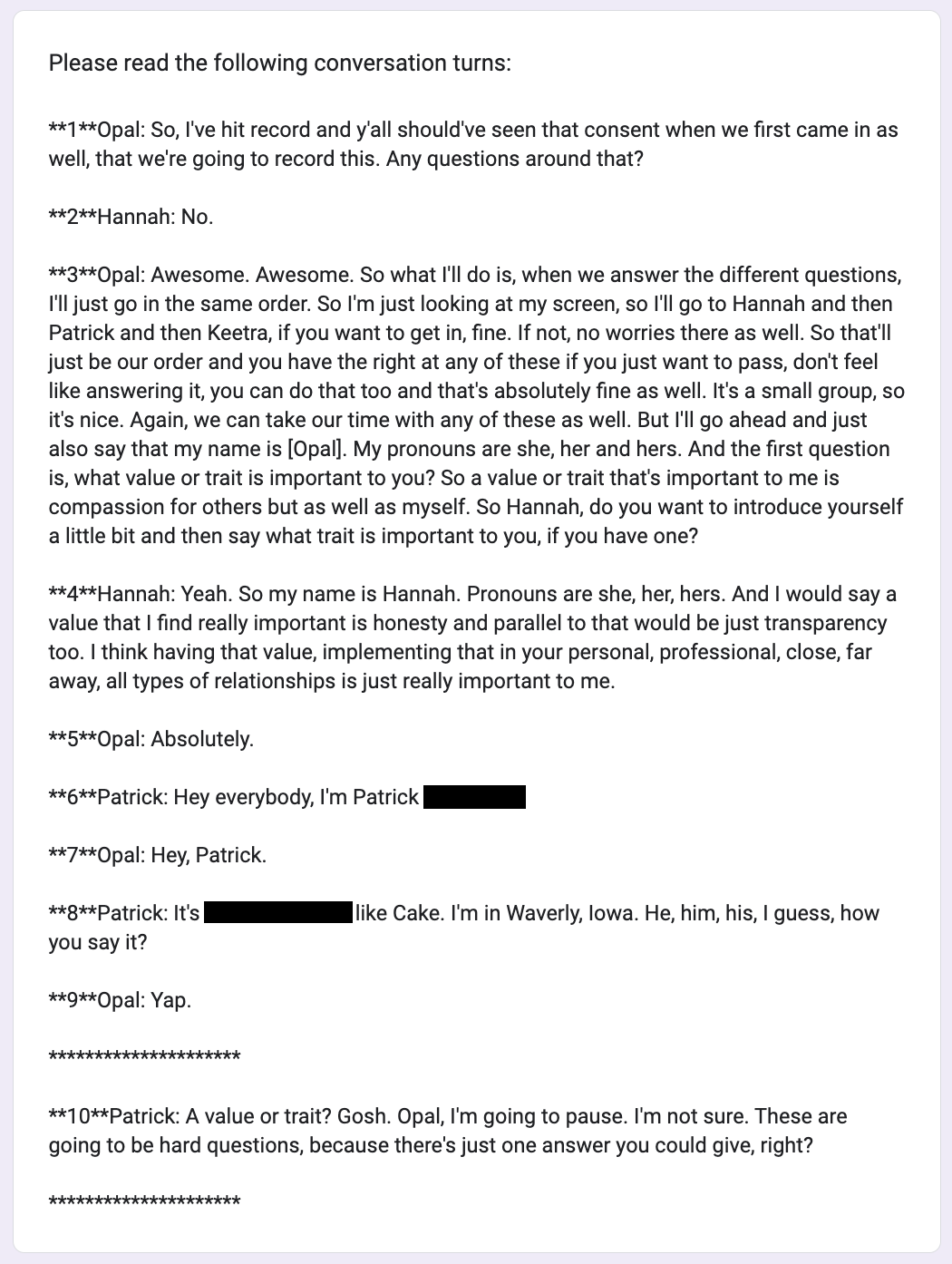}
    \label{fig:Form1}
\end{figure*}
\begin{figure*}
    \centering
    \includegraphics[width=1\linewidth]{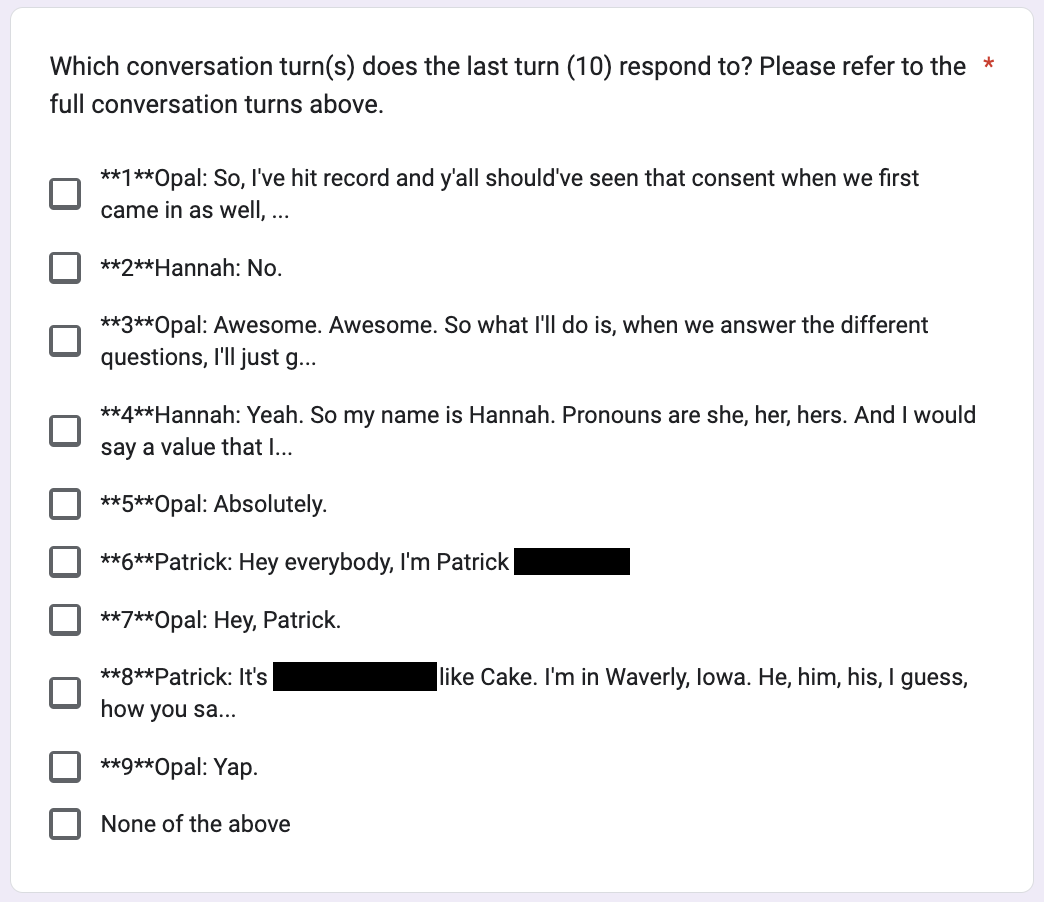}
    \caption{Form showing the human annotation task.}
    \label{fig:Form2}
\end{figure*}

\end{document}